\documentclass[runningheads]{llncs}
\usepackage{graphicx}
\usepackage{amsmath,amssymb} 
\usepackage[sort]{cite}

\begin{document}

\title{Single Image Super-Resolution Using Lightweight CNN with Maxout Units\thanks{This research was supported by Basic Science Research Program through the National Research Foundation of Korea (NRF) funded by the Ministry of  Science, ICT \& Future Planning (No. 2017R1A2A2A05001476).}} 
\titlerunning{SR Using Lightweight CNN with Maxout} 


\author{Jae-Seok Choi\inst{1} \and
Munchurl Kim\inst{1}}


\authorrunning{J.-S. Choi and M. Kim} 


\institute{School of EE, Korea Advanced Institute of Science and Technology, Korea\\
\email{\{jschoi14,mkimee\}@kaist.ac.kr}}

\maketitle

\begin{abstract}
Rectified linear units (ReLU) are well-known to obtain higher performance for deep-learning-based applications. However, networks with ReLU tend to perform poorly when the number of parameters is constrained. To overcome, we propose a novel network utilizing maxout units (MU), and show its effectiveness on super-resolution (SR). In this paper, we first reveal that MU can make the filter sizes halved in restoration problems thus leading to compaction of the network. To the best of our knowledge, we are the first to incorporate MU into SR applications and show promising results. In MU, feature maps from a previous convolutional layer are divided into two parts along channels, which are compared element-wise and only their max values are passed to a next layer. Along with interesting properties of MU to be analyzed, we further investigate other variants of MU. Our MU-based SR method reconstructs images with comparable quality compared to previous SR methods, even with smaller parameters.

\end{abstract}

\section{Introduction}
Super-resolution (SR) methods aim to reconstruct high-resolution (HR) image or video contents from their low-resolution (LR) versions. The SR problem is known to be highly ill-posed, where an LR input can lead to multiple degraded HR versions without proper prior information \cite{yang2010image}. As the role of SR becomes crucial recently in various areas such as up-scaling full-high-definition (FHD) to 4K \cite{choi2016super}, it is important to develop SR algorithms that are capable of generating HR contents with superior visual quality while maintaining reasonable complexity and moderate amounts of parameters.

\subsection{Related work}
SR methods can be divided into two families according to their input types: single image SR (SISR) and video SR. While both spatial and temporal information can be used in video SR, SISR utilizes only spatial information within given LR images, making the SR problem more difficult \cite{freedman2011image,shi2016real}. In this paper, we mainly focus on SISR.

\begin{figure}[t]
\centering
\includegraphics[width=0.7 \textwidth]{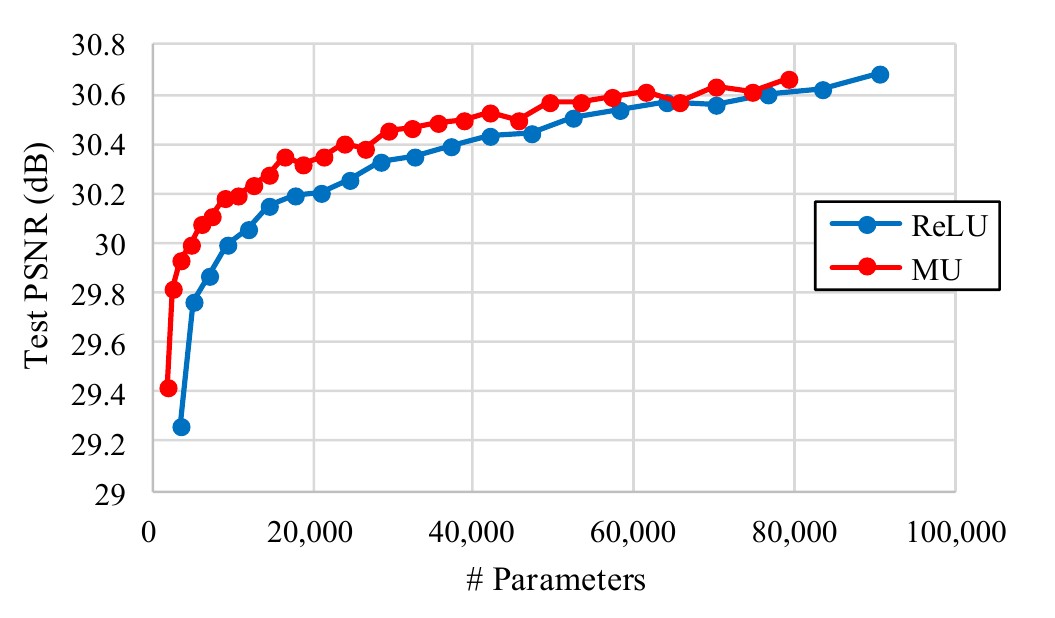}
\caption{Comparison on PSNR performance versus the number of filter parameters for two toy SR networks with ReLU \cite{nair2010rectified} and maxout units (MU) \cite{goodfellow2013maxout}, respectively. The network with MU shows higher performance than the conventional network with ReLU, especially when the number of parameter is small. This makes MU a suitable unit for application platforms with limited resources, such as mobile platforms.}
\label{fig:1}
\end{figure}

Various SR methods employed the following techniques in reconstructing HR images of high quality: sparse-representation \cite{jianchao2008image,kim2010single,yang2010image}, linear mappings \cite{choi2017single,timofte2014a2,zhang2012single,zhang2015learning,choi2016super}, self-examples \cite{freedman2011image,freeman2002example,glasner2009super,yang2010exploiting}, and neural networks \cite{choi2017deep,dong2014learning,dong2016image,kim2016accurate,ledig2017photo,lim2017enhanced,shi2016real,tai2017image,timofte2017ntire}. Sparse-representation-based SR methods \cite{jianchao2008image,kim2010single,yang2010image} undergo heavy computations to calculate sparse-representation of an LR patch from a pre-trained and complex LR dictionary. The resultant sparse-representation is then applied to a corresponding HR dictionary to reconstruct its HR version. Some SR methods \cite{freedman2011image,freeman2002example,glasner2009super,yang2010exploiting} extracted LR-to-HR mappings by searching for similar patches (self-examples) to the current patch inside its self-dictionary. Linear-mapping-based SR methods \cite{choi2016super,choi2017single,timofte2014a2,zhang2012single,zhang2015learning} (LMSR) have been proposed to obtain HR images of comparable quality but with much lower computational complexity. The adjusted anchored neighborhood regression (A+, APLUS) \cite{timofte2014a2} method searches for the best linear mapping for each LR patch, based on the correlation with pre-trained dictionary sets from \cite{yang2010image}. Choi \cite{choi2016super,choi2017single} employs simple edge classification to find suitable linear mappings, which are applied directly to small LR patches to reconstruct their HR version.

Recently, SR methods using convolutional neural networks (CNN) \cite{choi2017deep,dong2014learning,dong2016image,kim2016accurate,ledig2017photo,lim2017enhanced,shi2016real,tai2017image,timofte2017ntire} have shown high PSNR performance. Dong \textit{et al}. \cite{dong2014learning} first utilized a 3-layered CNN for SR (SRCNN), and reported a remarkable performance jump compared to previous SR methods. Recently, Kim \textit{et al}. \cite{kim2016accurate} proposed a very deep 20-layered CNN (VDSR) with gradient clipping and residual learning, yielding the reconstructed HR images of even higher PSNR compared to SRCNN. Shi \textit{et al}. \cite{shi2016real} proposed a network structure where features are extracted in LR space. The feature maps at the last layer are up-scaled to HR space using a sub-pixel convolution layer. Recursive convolutions were also used in \cite{tai2017image} to lower the number of parameters. Ledig \textit{et al}. \cite{ledig2017photo} presented two SR network structures: a network using residual units to maximize PSNR performance (SRResNet), and a network using generative adversarial networks for perceptual improvement (SRGAN). Lately, some SR methods using very deep networks \cite{choi2017deep,lim2017enhanced,timofte2017ntire} with large parameters have been proposed in NTIRE2017 Challenge \cite{timofte2017ntire}, achieving the state-of-the-art PSNR performance.

In these deep learning-based SR methods, rectified linear units (ReLU) \cite{nair2010rectified} are used to obtain nonlinearity between two adjacent convolutional layers. ReLU is a simple function, which has an identity mapping for positive values and 0 for negative. Unlike a sigmoid or hyperbolic tangent, ReLU does not suffer from gradient vanishing problems. By using ReLU, networks can learn piece-wise linear mappings between LR and HR images, which results in the mapping with high visual quality and faster training convergence. There are other nonlinear activation functions such as leaky ReLU (LReLU) \cite{maas2013rectifier}, parametric ReLU \cite{he2015delving} and exponential linear units (ELU) \cite{clevert2015fast}, but they are not often used in regression problems unlike ReLU. While LReLU replaces the zero part of ReLU with a linearity with certain small gradient, parametric ReLU parameterizes this gradient value so that a network can learn it. ELU has been designed so that it pushes mean unit activations closer to zero for faster learning.

\subsection{Motivations and contributions}

One major reason for such high performance of neural networks in many applications \cite{choi2017deep,dong2014learning,dong2016image,kim2016accurate,ledig2017photo,lim2017enhanced,shi2016real,tai2017image,timofte2017ntire} would be the use of ReLU \cite{nair2010rectified} and its successors \cite{maas2013rectifier}. These nonlinear units were first introduced in classification papers \cite{ba2016layer,clevert2015fast,he2015delving,ioffe2015batch,jia2014caffe,krizhevsky2012imagenet,maas2013rectifier,nair2010rectified}, which were subsequently reused for regression problems such as SR. It can be easily noticed that while ReLU and LReLU functions have been frequently used in SR, it is hard to find other types of activation functions \cite{clevert2015fast}. This is because they tend to distort scales of input values (more in Section 3.3), and thus networks with these functions generate HR results with lower quality compared to those with ReLU. This phenomenon can also be observed in normalization layers such as batch normalization \cite{ioffe2015batch} and layer normalization \cite{ba2016layer}, and there have been some reports that these normalization layers degrade performance when used in regression problems \cite{choi2017deep,lim2017enhanced}.

In this paper, we try to tackle some limitations of ReLU: i) ReLU produces feature maps with many zeros whose number is not controllable; ii) therefore, learning with ReLU tends to collapse in a network with very deep layers without some help such as identity mappings \cite{he2016identity}; and iii) there could be a way to make use of those empty zero values so that we may be able to reduce number of channels for lower memory consumption and less computations.

Maxout units (MU) \cite{goodfellow2013maxout} are activation units which could overcome the aforementioned limitations. MU were first introduced in various classification problems \cite{chang2015batch,goodfellow2013maxout,swietojanski2014investigation}. Goodfellow \textit{et al}. \cite{goodfellow2013maxout} proposed MU and used them in conjunction with dropout \cite{srivastava2014dropout} in a multi-layer-perceptron (MLP), and showed competitive classification results, compared to those of using conventional ReLU \cite{nair2010rectified}. In \cite{swietojanski2014investigation}, MU were used for speech recognition, and it is stated that networks with MU were about three times faster to converge in training with comparable performance. In addition, Chang \textit{et al}. \cite{chang2015batch} reported a network-in-network structure using MU for classification, which was able to mediate the problem of vanishing gradients that can occur when using ReLU. Although networks using MU were known to work well in high-level vision areas, only a few works \cite{cai2016dehazenet} employed MU for regression problems. In this paper, we develop and present a novel SR network incorporating MU. Our contributions are as follows:

\begin{itemize}
	\item Contrary to common thought that the number of parameters needs to be doubled when using MU, we first reveal that MU can effectively be incorporated into restoration problems. We show our SR network with MU that the number of channels of input feature maps is halved, even showing good results and thus resulting in a less memory usage and lower computational costs.
	\item We show a deep analysis on networks using basic MU, and further investigate other MU variants, showing their effectiveness on the SR application.
\end{itemize}

Various experiment results show that our SR networks that incorporate MU as activation functions are able to reconstruct HR images of competitive quality compared to those of ReLU. Figure \ref{fig:1} shows comparison on PSNR performance versus the number of parameters for two toy network examples with ReLU and MU, respectively. Both networks share the same 6-layered SR structure, except the type of activation functions used.

\section{Maxout units}

First, let us denote the outputs of the \textit{l}-th convolution layer as $\textbf{x}^l$, where a network has \textit{L} convolutional layers. Also, we denote the outputs of an activation function for $\textbf{x}^l$ as $\textbf{a}^l$.

\subsection{Conventional nonlinear activation functions}

Many SR methods \cite{choi2017deep,dong2014learning,dong2016image,kim2016accurate,ledig2017photo,lim2017enhanced,shi2016real,tai2017image,timofte2017ntire} often use ReLU \cite{nair2010rectified} for activation functions between every two convolutional layers to obtain high nonlinearity between LR and HR. After each ReLU, the negative part of feature maps $\textbf{x}^l$ becomes zero as

\begin{align}  
	\textbf{a}^{l} & = max(\textbf{x}^l,0),
\end{align}
where \textit{max}() is a function that calculates maximum values between two inputs in element-wise fashion. The negative parts where inputs become zero ensure nonlinearity, while the positive parts allow for fast learning as its derivative is a unity. However, very deep or narrow networks may have some difficulty in learning when too many values fall into negative and become zero. While other ReLU variants such as LReLU \cite{maas2013rectifier} and ELU \cite{clevert2015fast} try to overcome this limitation by modifying the negative parts, these ReLU variants still have little control over a ratio of the number of negative values.

\begin{figure}[t]
\centering
\includegraphics[width=1 \textwidth]{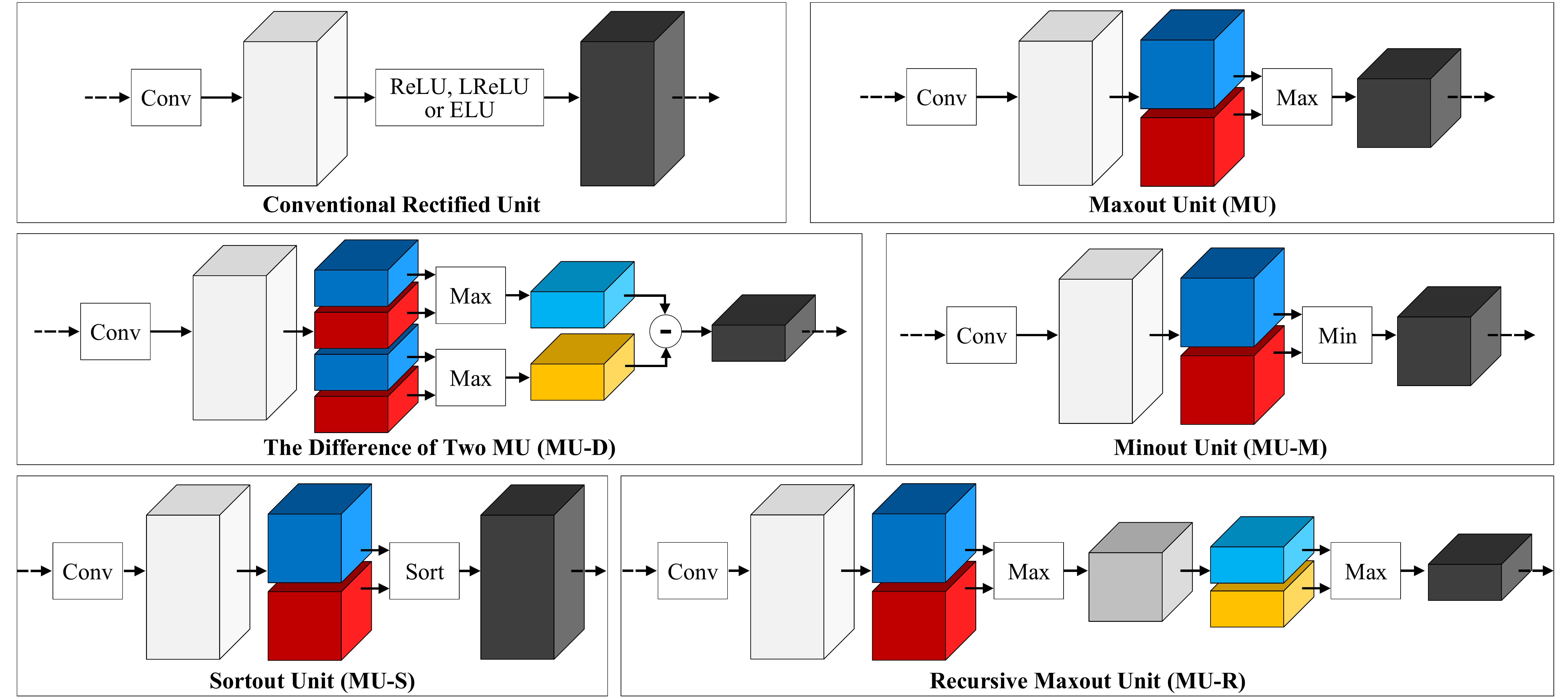}
\caption{Block diagrams for various activation functions used in networks: conventional units (ReLU \cite{nair2010rectified}, LReLU \cite{maas2013rectifier}, ELU \cite{clevert2015fast}), MU \cite{goodfellow2013maxout}, MU-D \cite{goodfellow2013maxout}, MU-M, MU-S and MU-R.}
\label{fig:2}
\end{figure}

\subsection{Maxout unit}
To overcome the limitations, we come up with an SR network structure incorporating the MU.
\subsubsection{Maxout.}
MU \cite{goodfellow2013maxout} computes the maximum of a vector of any length. Here, we use a special case of MU, where the feature maps $\textbf{x}^l$ are halved along channel into two parts $\textbf{x}_1^l$ and $\textbf{x}_2^l$, and element-wise maximum of these two parts is calculated as:

\begin{align}  
	\textbf{a}^{l} & = max(\textbf{x}_1^l,\textbf{x}_2^l).
\end{align}

\subsubsection{Difference of two MU.}
In \cite{goodfellow2013maxout}, a difference of two MU was also introduced with a proposition that any continuous piece-wise linear function can be expressed as a difference of two convex piece-wise linear functions. In this paper, we use the form of:

\begin{align}  
	\textbf{a}^{l} & = max(\textbf{x}_1^l,\textbf{x}_2^l)-max(\textbf{x}_3^l,\textbf{x}_4^l),
\end{align}
where $\textbf{x}^l$ is equally divided into four parts $\textbf{x}_1^l$, $\textbf{x}_2^l$, $\textbf{x}_3^l$ and $\textbf{x}_4^l$. Note after this activation function, the input feature maps are reduced to quarter. We denote this MU variant as MU-D. Incorporating a simple max function between two sets of feature maps provides nonlinearity with various properties as follow:

\begin{itemize}
	\item MU simply transfers feature map values from the input layer to the next, acting as the linear parts of ReLU. In backpropagation, error gradients simply flow to the selected values (maximum).
	\item Because MU does not consider negative or positive values unlike ReLU, outputs of MU would always have certain values, alleviating a chance of creating many close-to-zero values in feature maps and failing in learning.
	\item In narrow networks where the number of channels of feature maps is small, the MU allows for stable learning, while networks with ReLU may converge poorly.
	\item MU always ensures 50\% sparsity: that is, 50\% of larger values of the feature maps would always be selected and transmitted to the next layer, while the other 50\% of the feature maps are not used. In backpropagation, there would be always 50\% of paths alive for error gradients to be back-propagated.
	\item As the output of MU is only 50\% of the previous feature map values, the number of convolutional filter parameters in the next layer can be reduced by half, lowering both computation time and memory consumption. Similarly, unlike ReLU, MU is able to compress the given feature maps by stopping the transmission of close-to-zero values in the feature maps. In doing so, the network compactness is improved by preserving needed information.
\end{itemize}

We demonstrate the effectiveness of MU through various experiments in Section 3. Based on the properties of MU, we further investigate other variants of MU.

\subsection{MU variants}
From MU, its variants can be designed while preserving similar properties: minimum, recursive and sorting.

\subsubsection{Minimum.}
Instead of using the max function, one can design activation functions with the min function as

\begin{align}  
	\textbf{a}^{l} & = min(\textbf{x}_1^l,\textbf{x}_2^l),
\end{align}
where \textit{min}() is a function that calculates minimum values between two inputs in element-wise fashion. In training, this variant works similar to the original MU. We denote this MU variant as MU-M.

\subsubsection{Sorting.}
If we are to maintain the size of feature maps as ReLU does, we can employ both \textit{max} and \textit{min} functions into one activation function as

\begin{align}  
	\textbf{a}^{l} & = cat(max(\textbf{x}_1^l,\textbf{x}_2^l),min(\textbf{x}_1^l,\textbf{x}_2^l)),
\end{align}
where \textit{cat}() is a function that concatenates all inputs along channels. We denote this MU variant as MU-S.

\subsubsection{Recursive.}
By using MU recursively for \textit{n} times before applying convolutions in the next layer, we can further enforce more sparsity, e.g. 75\%, resulting reduced feature maps as outputs. This can be expressed as

\begin{align}  
	\textbf{a}^{l} & = f^n(\textbf{x}_1^l,\textbf{x}_2^l),
\end{align}
where $f^n$ indicates \textit{n}-times repeated MU, whose output channels are reduced by ${1/2}^n$. We denote this MU variant as MU-R.

Figure \ref{fig:2} illustrates the various activation functions, including MU and MU variants. Through additional experiments using the MU variants, we confirmed that networks with the variants could be trained well as shown in Section 3.

\begin{figure}[t]
\centering
\includegraphics[width=1 \textwidth]{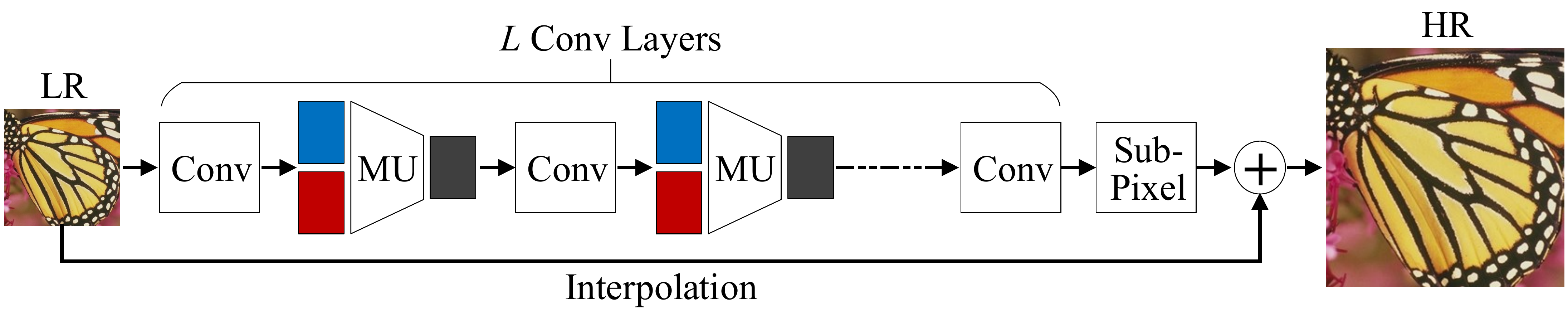}
\caption{Our proposed SR network, which incorporates MU for activation functions, residual learning between an interpolated image and a target HR image, and a sub-pixel convolution layer for up-scaling.}
\label{fig:3}
\end{figure}

\subsection{Network details}
By incorporating MU and its variants, we propose multiple network structures as shown in Figure \ref{fig:3}, and show their performance for SR applications.

\subsubsection{Toy networks.}
In order to conduct many and quick validations for comparing effects of multiple activation function variants including MU, we present a baseline toy network structure that is shared for testing all types of activation functions. The toy networks were trained using a smaller training dataset from \cite{yang2010image}. Our toy networks includes three types of layers: 6 layers of 3$\times$3 convolutions, one type of activation function, and one sub-pixel convolution layer \cite{shi2016real} at the end for up-scaling purpose. For convolutional layers, we simply use the kernel size of 3$\times$3, where input feature maps are padded with zero before convolution, so that the size of feature maps is preserved until the last sub-pixel convolution layer. The experimental results obtained using the toy networks are presented throughout Figures 1, 6, 7 and Table \ref{table:2}.

\subsubsection{ESPCN-MU.}
For comparison, several state-of-the-art SR network structures \cite{shi2016real,kim2016accurate} are implemented as stated in the papers but using MU and some modifications. Our first SR network using MU is based on ESPCN \cite{shi2016real}. We replace all ReLU layers in \cite{shi2016real} with MU. A 5-3-3 model \cite{shi2016real} is also used in our network with 64 filters for the first convolution layer and 32 filters for the second convolution layer. Note that due to MU’s characteristics where the number of channels is halved after activation, the number of filter parameters of ours is reduced almost in half compared to that of ESPCN \cite{shi2016real}. In addition, we aim to learn the residual between original HR images and interpolated LR images as in \cite{kim2016accurate}, but we use nearest-neighbor interpolation instead of bicubic to make SR problem harder and thus mainly focus on capability of types of activation functions. In doing so, networks converge faster. Due to its small number of parameters, we utilize a small training data set \cite{yang2010image}, but still produce comparable SR results to \cite{shi2016real}.

\begin{table}[t]
\centering
\caption{Average performance comparison for various SR methods.}
\label{table:1}
\renewcommand{\arraystretch}{1.5}
\scalebox{0.85}{
\begin{tabular} {c|c|c|c|c|c|c|c|c|c|c|c|c|c}
\hline \hline
\multicolumn{2}{c|}{Methods}&\multicolumn{2}{c|}{Bicubic}&\multicolumn{2}{c|}{SRCNN \cite{dong2016image}}&\multicolumn{2}{c|}{ESPCN \cite{shi2016real}}&\multicolumn{2}{c|}{ESPCN \cite{shi2016real}}&\multicolumn{2}{c|}{VDSR \cite{kim2016accurate}}&\multicolumn{2}{c}{SRResNet \cite{ledig2017photo}}\\
\hline
\multicolumn{2}{c|}{\# of Params}&\multicolumn{2}{c|}{-}&\multicolumn{2}{c|}{57K}&\multicolumn{2}{c|}{25K}&\multicolumn{2}{c|}{25K}&\multicolumn{2}{c|}{665K}&\multicolumn{2}{c}{923K}\\
\hline
\multicolumn{2}{c|}{Training Sets}&\multicolumn{2}{c|}{-}&\multicolumn{2}{c|}{ImageNet}&\multicolumn{2}{c|}{91}&\multicolumn{2}{c|}{ImageNet}&\multicolumn{2}{c|}{291}&\multicolumn{2}{c}{ImageNet}\\
\hline
Testing&Scale&PSNR&SSIM&PSNR&SSIM&PSNR&SSIM&PSNR&SSIM&PSNR&SSIM&PSNR&SSIM\\
\hline \hline
Set5&3&30.40&0.8687&32.75&0.9095&32.39&-&33.00&0.9121&33.66&0.9213&-&-\\
&4&28.43&0.8109&30.49&0.8634&-&-&30.76&0.8679&31.35&0.8838&32.06&0.8927\\
\hline \hline
Set14&3&27.55&0.7741&29.30&0.8219&28.97&-&29.51&0.8247&29.77&0.8314&-&-\\
&4&26.01&0.7023&27.50&0.7517&-&-&27.75&0.7580&28.01&0.7674&28.59&0.7811\\
\hline \hline
B100&3&27.21&0.7389&28.41&0.7867&-&-&-&-&28.82&0.7976&-&-\\
&4&25.96&0.6678&26.90&0.7107&-&-&-&-&27.29&0.7251&27.60&0.7361\\
\hline \hline
\multicolumn{14}{l}{*Results for the 9-5-5 model of SRCNN and results of ESPCN using ReLU are reported.}\\
\end{tabular}}

\medskip

\scalebox{0.85}{
\begin{tabular} {c|c|c|c|c|c|c|c}
\hline \hline
\multicolumn{2}{c|}{Methods}&\multicolumn{2}{c|}{\textbf{ESPCN-MU}}&\multicolumn{2}{c|}{\textbf{VDSR-MU}}&\multicolumn{2}{c}{\textbf{DNSR}}\\
\hline
\multicolumn{2}{c|}{\# of Params}&\multicolumn{2}{c|}{13K}&\multicolumn{2}{c|}{338K}&\multicolumn{2}{c}{133K}\\
\hline
\multicolumn{2}{c|}{Training Sets}&\multicolumn{2}{c|}{91}&\multicolumn{2}{c|}{291}&\multicolumn{2}{c}{291}\\
\hline
Testing&Scale&PSNR&SSIM&PSNR&SSIM&PSNR&SSIM\\
\hline \hline
Set5&3&32.85&0.9118&33.92&0.9231&33.80&0.9224\\
&4&30.57&0.8667&31.61&0.8861&31.57&0.8858\\
\hline \hline
Set14&3&29.40&0.8222&29.99&0.8346&29.95&0.8338\\
&4&27.61&0.7547&28.21&0.7713&28.21&0.7714\\
\hline \hline
B100&3&28.40&0.7853&28.87&0.7989&28.82&0.7980\\
&4&26.91&0.7114&27.31&0.7262&27.30&0.7260\\
\hline \hline
\end{tabular}}
\end{table}

\subsubsection{VDSR-MU.}
In addition, we propose another SR network using MU based on 20-layered VDSR \cite{kim2016accurate}. Similar to \cite{kim2016accurate}, 20 convolutional layers with 3$\times$3-sized filters are used in our network. We replace all ReLU layers in \cite{kim2016accurate} with MU. Similar to that of ESPCN-MU, the number of filters parameters of ours is reduced almost in half compared to that of VDSR \cite{kim2016accurate}. Also, we use nearest-neighbor interpolation instead of bicubic, and a sub-pixel convolution layer \cite{shi2016real} for faster computation speed. Due to its large number of parameters, our VDSR-MU network was trained using a larger data set combining \cite{yang2010image} and \cite{martin2001database} as in VDSR \cite{kim2016accurate}.

\subsubsection{DNSR.}
We also present a deeper and narrower version of VDSR-MU, called DNSR. While VDSR-MU has 20 layers with 64 channels, our DNSR has 30 layers (deeper) with 32 channels (narrower). Due to its deeper structure, we also employ residual units \cite{he2016identity} into DNSR for stable learning. Our DNSR holds a smaller number of total filter parameters, which is about 1/5 of that of VDSR \cite{kim2016accurate}, and about 1/2.6 of that of VDSR-MU, while showing PSNR performance similar to VDSR-MU.

\section{Experiment results}
We now demonstrate the effectiveness of MU and its variants in SR framework on popular image datasets, compared to conventional SR deep networks with common nonlinear activation functions, including ReLU.

\subsection{Experiment settings}
\subsubsection{Datasets.}
Two popular datasets \cite{yang2010image,martin2001database} were used for training networks. Images in the datasets were used as original HR images. Before given into networks, LR-HR training images are normalized between 0 and 1, and then LR training images are subtracted by 0.5 to have a zero mean. LR input images were created from these HR images by applying nearest-neighbor interpolation. SR process is only applied on Y-channel of YCbCr color space, and the chroma components, Cb and Cr, are up-scaled using simple bicubic interpolation. When comparing SR output images with original HR images, performance measures such as PSNR were done in Y-channel.

The training set of 91 images \cite{yang2010image} has frequently been used in various SR methods \cite{yang2010image,dong2014learning,kim2016accurate,shi2016real}. The dataset consists of small resolutions but with a variety of texture types. In our experiments, this smaller training set was used for various toy networks in order to conduct fast and many experiments, and was also used for our ESPCN-MU.

The Berkeley Segmentation Dataset \cite{martin2001database} has also been often used in SR works \cite{kim2016accurate,shi2016real}. This dataset includes 200 training images and 100 testing images for segmentation. As used in VDSR \cite{kim2016accurate}, we utilize 200 training images of BSD and 91 images from \cite{yang2010image} from training. This larger set was used for training VDSR-MU and DNSR.

For testing, three popular benchmark datasets including Set5 \cite{bevilacqua2012low}, Set14 \cite{zeyde2010single} and BSD100 \cite{martin2001database} were used.

\subsubsection{Training.}
We trained all the networks using ADAM \cite{kingma2014adam} optimization with an initial learning rate of $10^{-4}$ and the other hyper-parameters as defaults. We employed a uniform weight initialization technique in \cite{jia2014caffe} for training. All the networks including our proposed networks with MU were implemented using TensorFlow \cite{abadi2016tensorflow}, which is a deep learning toolbox for Python, and were trained/tested on GPU Nvidia Titan Xp.

The toy networks were trained for $10^{5}$ iterations, where a learning rate was lowered by a factor of 10 after $5\times10^{4}$ iterations. The mini-batch size was set to 2, weight decay was not used, and simple data augmentation with flip and rotation was used. For sub-images, LR-HR training image pairs were randomly cropped for the size of 40$\times$40 for a scale factor of 4.

Our ESPCN-MU, VDSR-MU and DNSR networks were trained for $10^{6}$ iterations, where a learning rate was lowered by a factor of 10 after $5\times10^{5}$ iterations. The mini-batch size was set to 4, and weight decay was not used. To create sub-images for training, LR-HR training image pairs were randomly cropped for the size of 75$\times$75 and 76$\times$76 in HR space, respectively, for a scale factor of 3 and 4. We apply various data augmentations to the HR images such as flipping, rotating, mirroring, and randomly multiplying their intensities by a value in a range from 0.8 and 1.2. Data augmentations are done on the fly for every epoch in training to reduce overfitting.

\subsection{SR results}
\begin{figure}[!t]
\centering
\includegraphics[width=1 \textwidth]{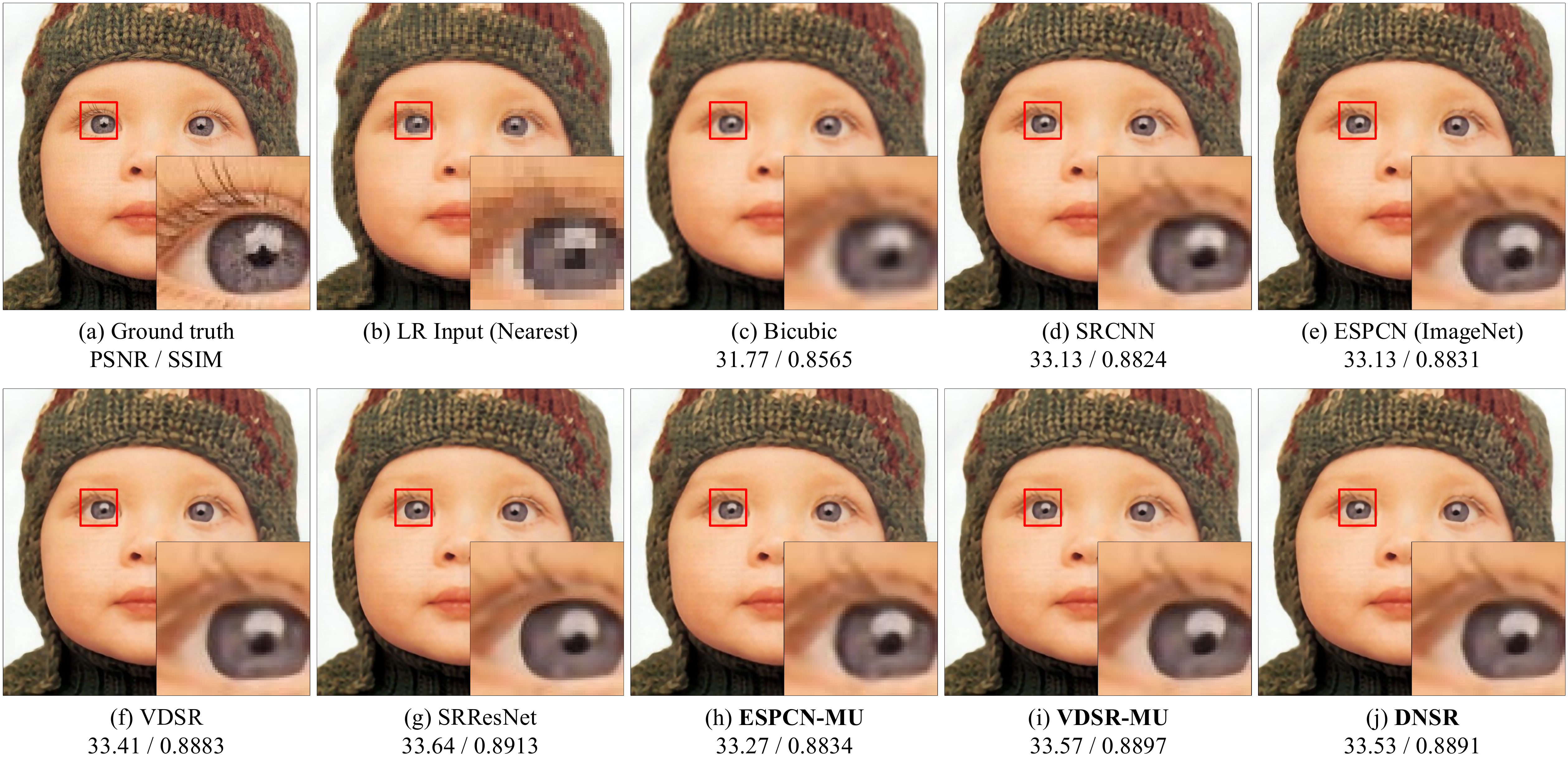}
\caption{Reconstructed HR images of \textit{baby} using various SR methods for a scale factor of 4.}
\label{fig:4}

\medskip

\includegraphics[width=1 \textwidth]{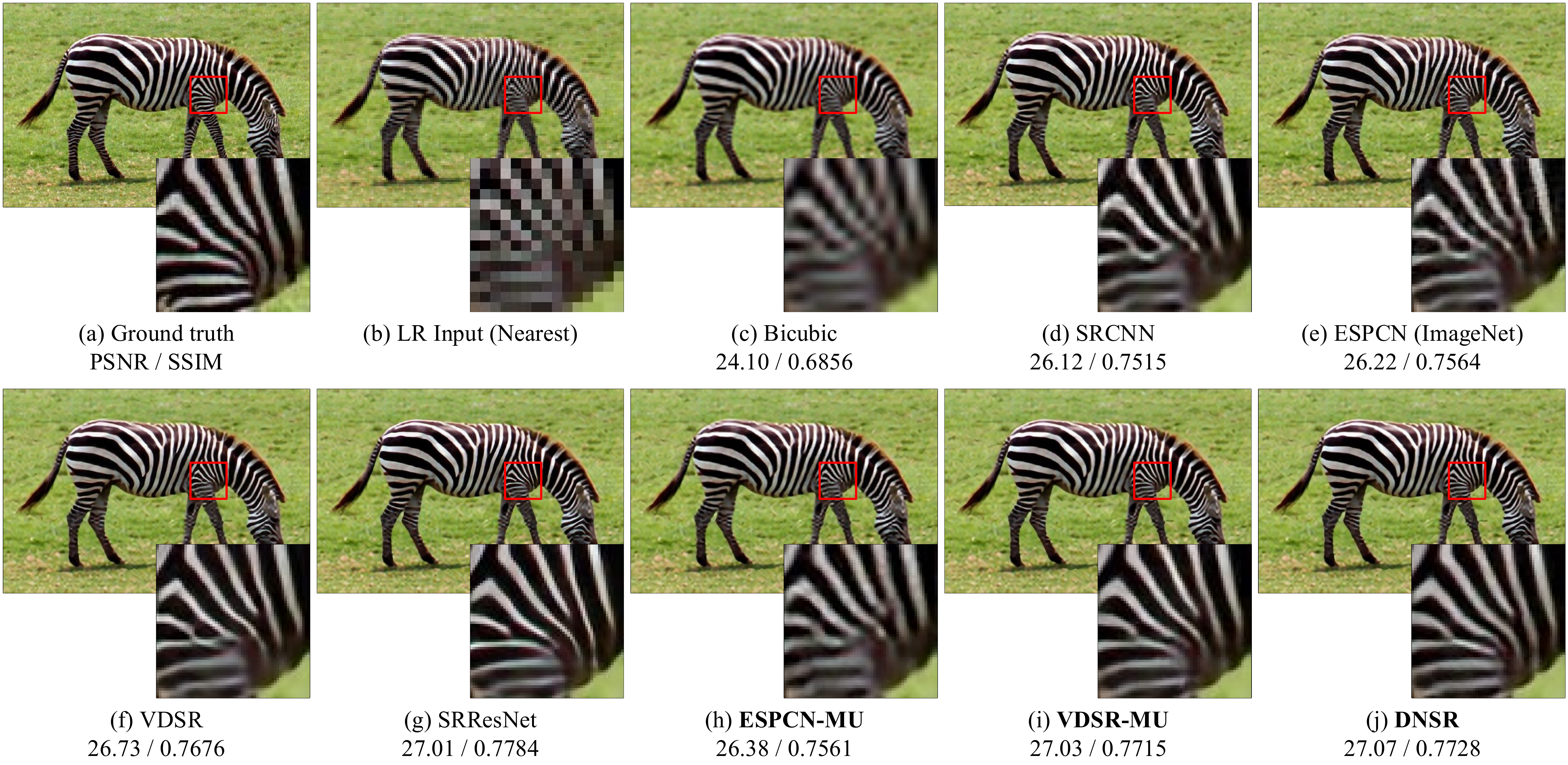}
\caption{Reconstructed HR images of \textit{zebra} using various SR methods for a scale factor of 4.}
\label{fig:5}
\end{figure}

First, we show SR results using our three proposed SR networks, including ESPCN-MU, VDSR-MU and DNSR, and compare them with the state-of-the-art methods, including SRCNN \cite{dong2016image}, ESPCN \cite{shi2016real}, VDSR \cite{kim2016accurate} and SRResNet \cite{ledig2017photo}. Table \ref{table:1} summarizes performance details for all the SR methods, including their numbers of filter parameters, their used training sets, and PSNR and SSIM \cite{wang2004image} values for scale factors of 3 and 4, tested on three popular testing datasets. For SRCNN \cite{dong2016image}, the reported results of the 9-5-5 model are shown. For ESPCN \cite{shi2016real}, the reported results using ReLU for two different training datasets are shown in Table \ref{table:1}. The PSNR/SSIM values for the conventional SR methods in Table \ref{table:1} are either the ones reported in their respective papers, or directly calculated from their publically available result images online. Figure \ref{fig:4} and \ref{fig:5} show reconstructed HR images and their magnified portions of \textit{baby} and \textit{zebra}, respectively, using various SR methods for a scale factor of 4.

\subsubsection{SR performance.}
As shown in Table \ref{table:1}, SRResNet \cite{ledig2017photo}, an SR network of the largest number of filter parameters (about 900K) that was trained using ImageNet \cite{russakovsky2015imagenet}, shows the highest PSNR and SSIM performance among various SR methods. Our proposed VDSR-MU and DNSR show the second and third highest performance with only 338K and 133K parameters, respectively, outperforming most of the conventional SR methods except SRResNet. It can be seen that our networks using MU have good efficiency with much less parameters, compared to other SR methods, while showing reasonable PSNR performance. As shown in Figure \ref{fig:4} and \ref{fig:5}, the quality of the reconstructed HR images using our VDSR-MU and DNSR are comparable to that of SRResNet \cite{ledig2017photo}. Especially, our VDSR-MU and DNSR were able to reconstruct clearly discerned stripes of \textit{zebra} as shown in Figure \ref{fig:5}-(i) and (j), which are comparable to Figure \ref{fig:5}-(g) of SRResNet, while other SR methods fail to do so.

\subsubsection{ESPCN-MU.}
In order to show the effectiveness of using MU in SR, we compare two similar networks: ESPCN \cite{shi2016real} and our ESPCN-MU. As shown in Table \ref{table:1}, the number of parameters of ESPCN-MU is only about 13K, which is almost half the number of parameters of ESPCN \cite{shi2016real}. While our network was trained using 91 images, our ESPCN-MU outperforms ESPCN \cite{shi2016real} trained with 91 training images, and shows comparable performance even compared to ESPCN \cite{shi2016real} that was trained using a larger set of images from ImageNet \cite{russakovsky2015imagenet}.

\begin{figure}[!t]
\centering
\includegraphics[width=0.6 \textwidth]{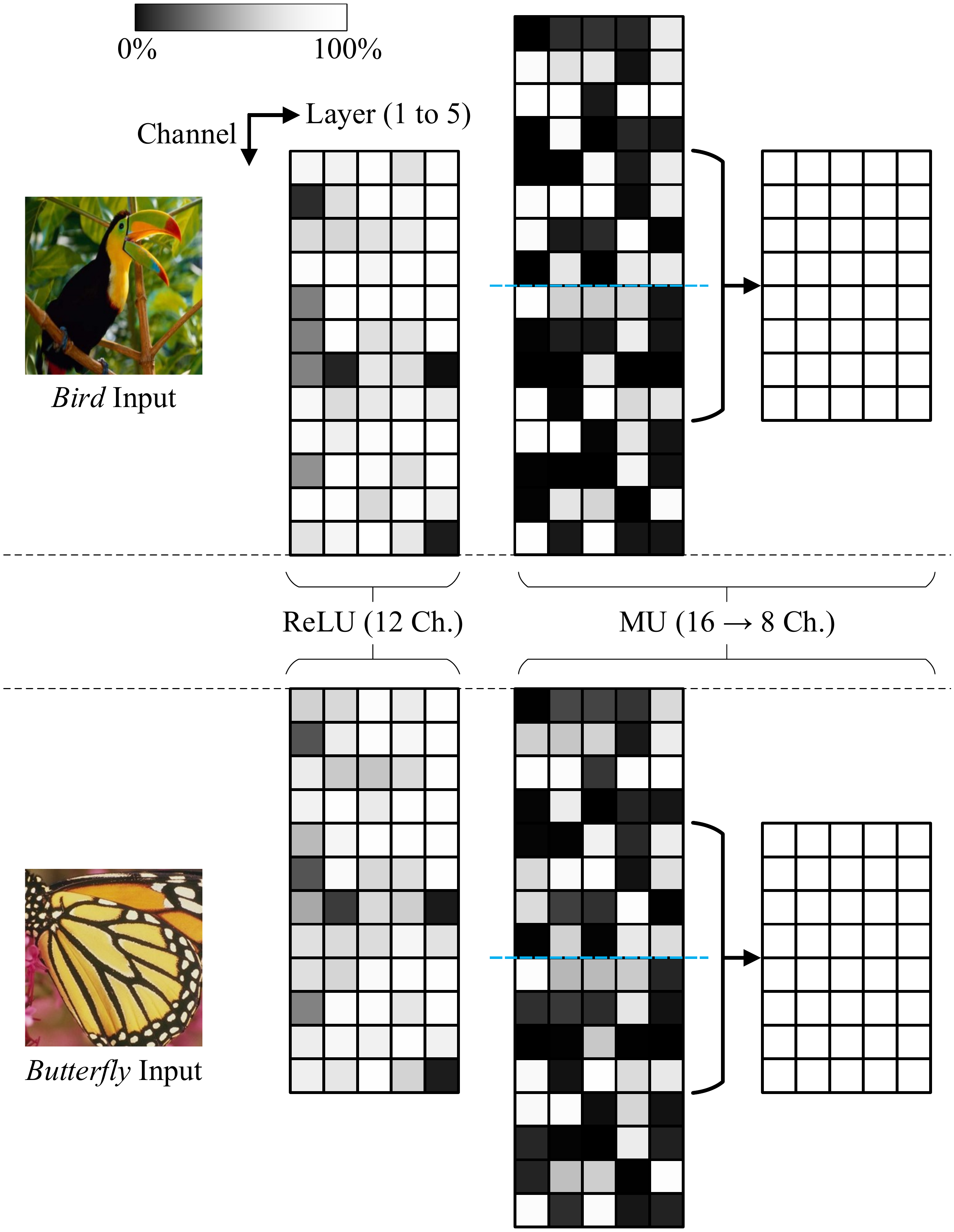}
\caption{Average ratios of activated neurons for each feature map in two toy networks using ReLU and MU for Bird and Butterfly, respectively (e.g. 100\% activation is colored in white, while 0\% in black and 50\% in gray). Here, rows and columns indicate channels and layers, respectively.}
\label{fig:6}
\end{figure}

\subsubsection{VDSR-MU.}
Similar to our ESPCN-MU, our VDSR-MU outperforms VDSR \cite{kim2016accurate} in terms of PSNR and SSIM, but with a much lower number of parameters. Our VDSR-MU networks has about 338K parameters, which is about half the number of parameters of VDSR \cite{kim2016accurate}. Note both networks were trained using the same 291 images \cite{yang2010image,martin2001database}.

\subsubsection{DNSR.}
Our deeper and narrower version of VDSR-MU, DNSR, has an almost 2/5 times the number of filter parameters compared to our VDSR-MU, which is about 1/5 of VDSR \cite{kim2016accurate} and 1/7 of SRResNet \cite{ledig2017photo}. Even with a low number of parameters, our DNSR network was able to reconstruct HR images comparable to VDSR-MU, and outperforms most of the conventional SR methods.

\begin{table}[!t]
\centering
\caption{Training and testing PSNR (dB) performance after the first $10^5$ iterations for networks with various activation functions.}
\label{table:2}
\renewcommand{\arraystretch}{1.2}
\scalebox{0.9}{
\begin{tabular}{c|c|c|c|c}
\hline \hline
Activation Function&Size of Conv Filters&Number of Params&Training PSNR&Testing PSNR\\
\hline
ReLU \cite{nair2010rectified}&$3\times3\times12\times12$&7.1K&28.22&29.81\\
LReLU \cite{maas2013rectifier}&$3\times3\times12\times12$&7.1K&28.19&29.78\\
ELU \cite{clevert2015fast}&$3\times3\times12\times12$&7.1K&27.91&29.30\\
\hline
MU \cite{goodfellow2013maxout}&$3\times3\times8\times16$&6K&28.42&30.07\\
MU-D \cite{goodfellow2013maxout}&$3\times3\times6\times24$&6.4K&28.46&30.07\\
\hline
MU-M&$3\times3\times8\times16$&6K&28.43&30.05\\
MU-S&$3\times3\times12\times12$&7.1K&28.46&30.08\\
MU-R&$3\times3\times6\times24$&6.4K&28.38&29.98\\
\hline \hline
\end{tabular}}
\end{table}

\subsection{Discussions}
We also conducted experiments on toy networks using various activation functions including ReLU, MU and MU variants. We show potential properties of MU compared to units used in conventional SR methods, by analyzing parameter-vs.-PSNR performance and by showing activation rates in feature maps.

\subsubsection{MU versus ReLU.}
Figure \ref{fig:1} shows comparison on PSNR performance versus the number of parameters for two toy networks with ReLU and MU, respectively. Both networks share the same 6-layered SR structure, except the type of activation functions used. The number of parameters for each subtest is controlled by adjusting the number of convolution filters. As shown, the PSNR performance gap between networks using ReLU and MU becomes larger as the number of parameters decreases. This indicates that in narrow networks where the number of channels of feature maps is small, MU allows for stable learning, while ReLU converges towards a worse point. We can argue that because MU does not consider negative or positive values unlike ReLU, the outputs of MU would always have certain values, alleviating a chance of creating many close-to-zero values in feature maps and failing in learning.

Figure \ref{fig:6} shows the average ratios of activated neurons for each feature map in two toy networks using ReLU and MU for \textit{Bird} and \textit{Butterfly}, respectively (e.g. 100\% activation is colored in white, 0\% in black and 50\% in gray). The rows and columns indicate channels and layers, respectively. It is interesting to see that activations after MU are sparser than those of ReLU, which supports the effectiveness of MU in SR. Note that since the maximum values between two feature maps are always passed to the next layer, the feature maps after MU would always be 100\% activated with half the number of feature maps. Note that Figure \ref{fig:6} may suggest that MU can be related to network pruning, and this remains as our future work.

\subsubsection{MU variants.}
\begin{figure}[!t]
\centering
\includegraphics[width=0.7 \textwidth]{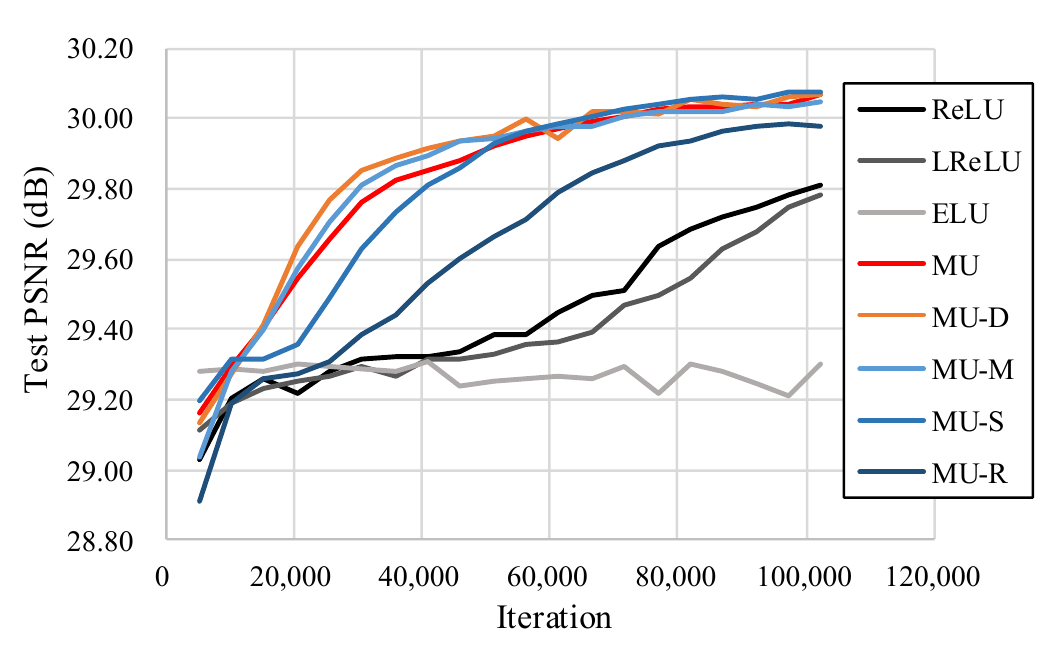}
\caption{A PSNR-versus-iteration plot for networks with various activation functions from Table \ref{table:2}.}
\label{fig:7}
\end{figure}

Table \ref{table:2} shows training and testing PSNR performance after the first $10^{5}$ iterations for toy networks with various activation functions. Note that the size of convolutional filters has been adjusted for each network to yield a similar number of total parameters. Figure \ref{fig:7} presents a PSNR-versus-iteration plot for networks with various activation functions from Table \ref{table:2}. It can be seen in Figure \ref{fig:7} that the networks with MU and MU variants enable faster convergence compared to those with ReLU and ReLU variants. Note that the network with ELU has a training difficulty in the SR problem, contrary to its performance in other classification papers. This may be due to the fact that ELU tend to distort scales of input values, which is undesirable in regression problems such as SR. Overall, the networks with MU and its variants show higher PSNR values with less parameters.

\section{Conclusion}
The proposed SR networks showed superior PSNR performance compared to the base networks using ReLU and other activation functions. The SR networks using MU tend to produce higher PSNR results with a smaller number of convolution filter parameters, which is desirable for computational platforms with limited resources. We showed that MU can be used in regression problems especially SR, and they have some potential with further extension to new types of activation functions for other applications. 

\clearpage

\bibliographystyle{splncs04}
\bibliography{egbib}
\end{document}